\relax
\documentclass[letterpaper]{article} % DO NOT CHANGE THIS
\usepackage{aaai20}  % DO NOT CHANGE THIS
\usepackage{times}  % DO NOT CHANGE THIS
\usepackage{helvet} % DO NOT CHANGE THIS
\usepackage{courier}  % DO NOT CHANGE THIS
\usepackage[hyphens]{url}  % DO NOT CHANGE THIS
\usepackage{graphicx} % DO NOT CHANGE THIS
\usepackage{natbib}
\usepackage{graphicx}  % DO NOT CHANGE THIS
\usepackage{enumitem}
\title{On Measuring and Mitigating Biased Inferences of Word Embeddings}
\author{\\ \Large \textbf{Sunipa Dev, Tao Li, Jeff M. Phillips, Vivek Srikumar}\\ % All authors must be in the same font size and format. Use \Large and \textbf to achieve this result when breaking a line
School of Computing\\ 
University of Utah\\
Salt Lake City, Utah, USA\\%If you have multiple authors and multiple affiliations
\{sunipad, tli, jeffp, svivek\}@cs.utah.edu 
}
 
\begin{document}

\maketitle

\begin{abstract}
Word embeddings carry stereotypical connotations from the text they
are trained on, which can lead to invalid inferences in downstream models that
rely on them.  
We use this observation to design a mechanism for measuring stereotypes using the task of natural language inference.  
We demonstrate a reduction in invalid inferences via bias mitigation strategies on static word embeddings (GloVe). Further, we show that for gender bias, these techniques extend to contextualized embeddings when applied selectively only to the static components of contextualized embeddings (ELMo, BERT).
\end{abstract}

\section{Introduction}
\label{sec:intro}
Word embeddings have become the de facto feature representation
across
NLP~\cite[for example]{parikh2016decomposable,seo2016bidirectional}.
Their usefulness stems from their ability capture background
information about words using large corpora as static vector
embeddings---e.g., word2vec~\cite{Mik1}, GloVe~\cite{Art3}---or contextual
encoders that produce embeddings---e.g., ELMo~\cite{Peters:2018}, BERT~\cite{bert}.

However, besides capturing word meaning, their embeddings
also encode real-world biases about gender, age, ethnicity, etc. To
discover biases, several lines of existing
work~\cite{debias,Caliskan183,ZhaoWYOC17,Bias1} employ
measurements intrinsic to the vector representations, which despite
their utility, have two key problems. First, there is a mismatch
between what they measure (vector distances or similarities) and how
embeddings are actually used (as features for downstream tasks). Second, contextualized embeddings like ELMo or BERT drive today's state-of-the-art NLP systems, but tests
for bias are designed for word types, not
word \emph{token} embeddings.

In this paper, we present a general strategy to probe word embeddings for
biases.  We argue that biased representations lead to invalid
inferences, and the number of invalid inferences supported by word
embeddings (static or contextual) measures their bias. To concretize
this intuition, we use the task of natural language inference (NLI),
where the goal is to ascertain if one sentence---the premise---\emph{entails} or \emph{contradicts} another---the hypothesis, or if
neither conclusions hold (i.e., they are \emph{neutral} with respect
to each other).
 
 \vspace{2cm}
As an illustration, consider the sentences:
\begin{enumerate} [label=(\arabic*), leftmargin = 1.7em]
	\item \label{ex:rude-bishop1} The rude person visited the bishop.
	\item \label{ex:uzbek-bishop1} The Uzbekistani person visited the bishop.
\end{enumerate}

Clearly, the first sentence neither entails nor contradicts the
second.  Yet, the popular decomposable attention
model~\cite{parikh2016decomposable} built with GloVe embeddings
predicts that sentence \ref{ex:rude-bishop1} entails sentence
\ref{ex:uzbek-bishop1} with a high probability of $0.842$!  Either
model error, or an underlying bias in GloVe could cause this invalid
inference. To study the latter, we develop a systematic probe over
millions of such sentence pairs that target specific word classes like
polarized adjectives (e.g., rude) and demonyms (e.g., Uzbekistani).

A second focus of this paper is bias attenuation. As a
representative of several lines of work in this direction, we use
the recently proposed projection method of~\cite{Bias1}, which
identifies the dominant direction defining a bias
(e.g., gender), and removes it from \emph{all} embedded vectors.
This simple approach thus, avoids the trap of
residual information~\cite{gonen2019lipstick} seen in hard
debiasing approach of~\cite{debias}, which categorizes words and treats each category differently.  
Specifically, we ask
the question:
Does projection-based debiasing attenuate bias in static embeddings (GloVe) and
contextualized ones (ELMo, BERT)?

\paragraph{Our contributions.}
\emph{Our primary contribution is the use of natural language
	inference-driven to design probes that measure the effect of specific biases.} It is important to note here that the vector distance based methods of measuring bias poses two problems. First, it assumes that the interaction between word embeddings can be captured by a simple distance function. Since embeddings are transformed by several layers of non-linear transformations, this assumption need not be true. Second, the vector distance method is not applicable to contextual embeddings because there is no single ‘driver’, ‘male’, ‘female’ vectors; instead the vectors are dependent on the context. Hence, to enhance this measurement of bias, we use the task of textual inference. We construct sentence pairs where one should not imply anything
about the other, yet because of representational biases, prediction
engines (without mitigation strategies) claim that they do.
To quantify this we use model probabilities for entailment (E),
contradiction (C) or neutral association (N) for pairs of sentences.
Consider, for example,

\begin{enumerate} [label=(\arabic*), leftmargin = 1.7em]  \setcounter{enumi}{2}
	\item \label{ex:driver-cabinet} The driver owns a cabinet.
	\item \label{ex:man-cabinet} The man owns a cabinet.
	\item \label{ex:woman-cabinet} The woman owns a cabinet.
\end{enumerate}

The sentence \ref{ex:driver-cabinet} neither entails nor
contradicts sentences \ref{ex:man-cabinet} and
\ref{ex:woman-cabinet}. Yet, with sentence
\ref{ex:driver-cabinet} as premise and sentence
\ref{ex:man-cabinet} as hypothesis, the decomposable attention
model predicts probabilities: 
E: 0.497, N: 0.238, C: 0.264; the model predicts entailment.
Whereas, with sentence \ref{ex:driver-cabinet} as premise and sentence \ref{ex:woman-cabinet} as hypothesis, we get  E: 0.040, N: 0.306, C: 0.654; the model predicts contradiction. 
Each premise-hypothesis pair differs only by a gendered word.

% To show the general applicability of this scheme, we
% expand on this idea, and build a suite of tasks also capturing
% representational bias in nationality or religion
% For instances, consider the following two sentences:
% \begin{exe}
% 	\ex \label{ex:rude-bishop} The rude person visited the bishop.
% 	\ex \label{ex:uzbek-bishop} The Uzbekistani person visited the bishop.
% \end{exe}
% A inference models built on top of GloVe predicts that sentence (\ref{ex:rude-bishop}) entails sentence (\ref{ex:uzbek-bishop}) with a high probability: $0.842$.  This should be neither an entailment or a contradiction, it should be neutral; hence this inference is not meaningful and instead represents societal bias.  

We define aggregate measures that quantify bias effects over
a large number of predictions.  
We discover substantial bias across GloVe,  ELMo and BERT embeddings. In
addition to the now commonly reported gender bias~\cite[for example]{debias}, we also show that the embeddings encode polarized information about demonyms and religions.
To our knowledge, this is the among the first demonstrations ~\cite{sweeney_transparent_2019,manzini-etal-2019-black} of national or religious bias in word embeddings.

%\vs{Perhaps the intro may be a good place to seed the idea that bias need not be just gender bias. So maybe choose examples that are not gender related, but more diffused. This will set up the question of how to measure bias.}

%Further, we extend these methods of bias detection and removal to newer context sensitive word embeddings like ELMo. In this word embedding framework, each occurrence of a word has a distinct position in the vector space, thus making the context a word occurred in pivotal in its vector encoding.

%ELMo encodings are deep contextualized representations of words made from one layer of stationary, context-insensitive token embedding and two layers of context sensitive bi directional LSTMS. The final representation of a word is a combination of these three encodings in the form of a concatenation or a linear interpolation between the three layers. Thus, the final representation of each token is context sensitive and a word that occurs $>1$ times has $>1$ vector representations. This makes it trickier to detect bias and subsequently remove or dampen it in the representation. 

%\vs{This paragraph is too much information for the intro}

\emph{Our second contribution is to show that simple mechanisms for removing bias on static word embeddings (particularly GloVe) work.}  The projection approach of~\cite{Bias1} has been shown effective for intrinsic measures; we show that its effectiveness extends to the new NLI-based probes.  Specifically, we show that it reduces gender's effect on occupations. We further show similar results for removing subspaces associated with religions and demonyms.

\emph{Our third contribution is that these approaches can be extended to contextualized embeddings (on ELMo and BERT), but with limitations.}  
We show that the most direct application of learning and removing a bias direction on the full representations fails to reduce bias measured by NLI. However, learning and removing a gender direction from the \emph{non-contextual part} of the representation (the first layer in ELMo, and subword embeddings in BERT), can reduce NLI-measured gender bias.  
Yet, this approach is ineffective or inapplicable for religion or nationality.

%many general approaches fail to reduce bias measured by NLI.
%For tasks involving gender bias, we show learning and removing a
%gender direction on all the three layers in an ELMo embedding does not
%reduce bias.  However, surprisingly, removing it from only the first
%layer (effectively a static word embedding) is effective for
%reducing gender bias.  Yet, this approach is ineffective for
%religion or nationality.

%On a positive note, the amount of gender bias ELMo encodes is roughly the same as GloVe after attenuating bias in both.  In the case of biases associated with religions or nationality, less bias is reported using a bias-attenuated GloVe than with ELMo either before or after attempts at reducing bias.  
%As such it seems 
%\emph{contextual embeddings may offer better predictive performance, but also seem to encode more and harder to remove biases.}

\section{Measuring Bias with Inference}
\label{sec:NLI}

Our construction of a bias measure
uses the NLI task, which has been widely studied in NLP, starting
with the PASCAL RTE
challenges~\cite{dagan2006the-pascal,dagan2013recognizing}. More
recently, research in this task has been revitalized by large
labeled corpora such as the Stanford NLI corpus~\cite[SNLI]{snli}.

The motivating principle of NLI is that inferring relationships
between sentences is a surrogate for the ability to reason about
text.  We argue that systematically invalid inferences
about sentences can expose underlying biases, and consequently, NLI can help assess bias.  We will describe this process
using how gender biases affect inferences related to occupations.
Afterwards, we will extend the approach to polarized inferences
related to nationalities and religions.

\subsection{Experimental Setup}
\label{sec:experimental-setup}
%Before describing our exploration of bias in representations, let us look at the word embeddings we study and the NLI systems trained over them.
%
We use GloVe to study static word embeddings and ELMo and BERT for contextualized ones.
Our NLI models for GloVe and ELMo are based on the decomposable attention model~\cite{parikh2016decomposable} with a BiLSTM encoder instead of the original projective one ~\cite{cheng2016long}. For BERT, we use BERT$_{\textrm{BASE}}$, and follow the NLI setup in the original work. 
Our models are trained on the SNLI training set. We list other details of our experiments here.

\paragraph{Static embeddings.} For static embeddings, we adopted the original DAN architecture but replaced the projective encoder with a bidirectional LSTM~\cite{cheng2016long} encoder.
We used the GloVe pretrained on the common crawl dataset with dimension $300$.
Across the network, the dimension of hidden layers are all set to $200$.
That is, word embeddings get downsampled to $200$ by the LSTM encoder.
Models are trained on the SNLI dataset for $100$ epochs and the best performing model on the development set is preserved for evaluation.

\paragraph{Context-dependent embeddings.} For ELMo, we used the same architecture as above except that we replaced the static embeddings with the weighted summation of three layers of ELMo embeddings, each 1024 dimensional vectors.
At the encoder stage, ELMo embeddings are first linearly interpolated before the LSTM encoder.
Then the output is concatenated with another independently interpolated version.
The LSTM encoder still uses hidden size $200$.
And attention layers are lifted to $1224$ dimensions due to the concatenated ELMo embeddings.
For classification layers, we extend the dimension to $400$.
Models are trained on the SNLI dataset for $75$ epochs.

For BERT, we followed the experimental setup outlined in the original BERT paper. Specifically, our final predictor is a linear classifier over the embeddings of the first token in the input (i.e, \texttt{[CLS]}).
Across our experiments, we used the pre-trained BERT$_{BASE}$ to further finetune on the SNLI dataset with learning rate $0.00003$ for $3$ epochs. During training, we used dropout $0.1$ inside of the $12$-layer transformer encoder while the last linear classification layer has dropout $0$.

\paragraph{Debiasing \& retraining.}
To debias GloVe, we removed corresponding components off the static embeddings of all words, using the  projection mechanism described in the main body of the paper.
The resulting embeddings are then used for (re)training.
To debias ELMo, we conduct the same removal method on the input character-based word embeddings, and then embed as usual.
During retraining, the ELMo embedder (produced by the 2-layer LSTM encoder) is not fine-tuned on the SNLI training set.
For our BERT experiments, we debiased the word piece (i.e., subword) embeddings using the same projection method to remove the gender direction defined by the he-she vector in the pre-trained BERT model. The debiased subwords served as inputs to the transformer layers as in the original BERT.

%\footnote{We will release the code for our experiments with the final version.}

%
%We re-implemented the decomposable attention model of
%\citet{parikh2016decomposable}. We replaced the projective encoder
%in the original paper with a BiLSTM over the word embeddings we wish
%to probe.  We use GloVe to study static word embeddings and ELMo for
%contextual ones. For ELMo, we linearly interpolated the three layers
%it produces to obtain inputs to the BiLSTM. The interpolation terms
%are trained with the rest of the parameters.  We trained GloVe and
%ELMo models for $100$ and $75$ epochs respectively. Our models are
%trained on the SNLI training set, and evaluated on the validation
%set at each epoch.  The best performing model is selected for final
%evaluation.  The appendix lists all hyper-parameters and network
%details.\footnote{We will publish all trained models upon
%  publication. We will also release code that not only reproduces
%  the measurements in this paper, but can also be used to probe for
%  new types of biases using this framework.}

\subsection{Occupations and Genders}
\label{sec:occupations-and-genders}

Consider the following three sentences:
\begin{enumerate} [label=(\arabic*), leftmargin = 1.7em] \setcounter{enumi}{5}
	\item \label{ex:accountant-bagel} The accountant ate a bagel.
	\item \label{ex:man-bagel} The man ate a bagel.
	\item \label{ex:woman-bagel} The woman ate a bagel.
\end{enumerate}
The sentence \ref{ex:accountant-bagel} should neither entail nor
contradict the sentences \ref{ex:man-bagel} and
\ref{ex:woman-bagel}: we do not know the gender of the accountant.
For these, and many other sentence pairs, the correct
label should be neutral, with prediction probabilities
E: 0, N: 1, C: 0. But a gender-biased representation of the word
\textsf{accountant} may lead to a non-neutral prediction.
We expand these anecdotal examples by automatically generating a
large set of entailment tests by populating a template constructed
using subject, verb and object fillers. All our templates are of the
form:
\begin{quote}
	The \textsf{subject} \textsf{verb} a/an \textsf{object}.
\end{quote}
Here, we use a set of common activities for the \textsf{verb} and
\textsf{object} slots, such as \textsf{ate a bagel},
\textsf{bought a car}, etc. For the same \textsf{verb} and
\textsf{object}, we construct an entailment pair using subject
fillers from sets of words. For example, to assess gender bias
associated with occupations, the premise of the entailment pair
would be an \textsf{occupation} word, while the hypothesis would be
a \textsf{gendered} word. The extended version of the paper has all the word lists we
use in our experiments.

Only the \textsf{subject} changes between the premise and the hypothesis in any pair.
Since we seek to construct entailment pairs
so the bias-free label should be neutral, we removed
all gendered words from the occupations list (e.g., \textsf{nun},
\textsf{salesman} and \textsf{saleswoman}).  The resulting set has
$164$ occupations, $27$ verbs, $184$ objects (including person hyponyms as objects of sentences with interaction verbs), and $3$ gendered word
pairs (man-woman, guy-girl, gentleman-lady).  Expanding these
templates gives us %$164 \cdot 27 \cdot 95 \cdot 6 = 2{,}523{,}960$
over $2,000,000$
entailment pairs, \emph{all} of which we expect are neutral. The extended version has the word lists, templates and other experimental details;  the code for template generation and  experimental setup is also online \footnote{https://github.com/sunipa/On-Measuring-and-Mitigating -Biased-Inferences-of-Word-Embeddings}.

% We consider XX occupations, and YY actions, listed in Appendix \ref{app:templates}.  
% Then each template consists of 3 sentences:
% \begin{exe}
% \ex The $o$ $a$.
% \ex The man $a$.
% \ex The woman $a$.
% \end{exe}
% And we generate two ENC vectors using ``The $o_i$ $a_i$." as the proposition and both ``The man $a_i$." and ``The woman $a_i$.'' as the hypothesis.  In total this results in $M = 2 \times  XX \times YY = ...$ ENC vectors. 

%%%%%%%%%%%%%%%%%%%%%%%%%%%%%%%%%%%%%%%%%%%%%%%%%%%%%%%%%%%%%%%%%%%%%
\subsection{Measuring Bias via Invalid Inferences}
\label{sec:metrics}
Suppose we have a large collection of $M$ entailment pairs $S$ constructed by populating templates as described
above. Since each sentence pair $s_i \in S$ should be inherently
neutral, we can define bias as deviation from neutrality.
Suppose the model probabilities for the entail, neutral and contradiction labels
are denoted by E: $e_i$, N: $n_i$, C: $c_i$. We define three different
measures for how far they are from neutral:
\begin{enumerate}
	\item \textbf{Net Neutral (NN)}: The average probability of
	the neutral label across all sentence pairs.
	$NN=\frac{1}{M} \sum_{i =1}^M n_i$.
	
	\item \textbf{Fraction Neutral (FN)}: The fraction of
	sentence pairs labeled neutral.
	$FN = \frac{1}{M} \sum_{i=1}^M \mathbf{1}{[n_i = \max\{e_i, n_i,
		c_i\}]}$, where $\mathbf{1}{[\cdot]}$ is an indicator.
	
	\item \textbf{Threshold:$\tau$ (T:$\tau$)}: A parameterized measure
	that reports the fraction of examples whose probability of neutral
	above $\tau$: we report this for $\tau = 0.5$ and $\tau = 0.7$.
\end{enumerate}
In the ideal (i.e., bias-free) case, all three measures will take the
value $1$.

\begin{table}[h!]
	\centering
	\begin{tabular}{rcccc}
		%\hline
		Embedding & NN  &  FN & T:$0.5$ & T:$0.7$ 
		\\ \hline
		GloVe & 0.387 & 0.394 & 0.324 & 0.114 
		\\
		ELMo & 0.417 & 0.391 & 0.303 & 0.063
		\\
		BERT  & 0.421 & 0.397 & 0.374 & 0.209
		\\ \hline
	\end{tabular}
	\caption{\label{tbl:Gender}
		Gender-occupation neutrality scores, for models using GloVe, ELMo, and BERT embeddings.}
\end{table}

Table \ref{tbl:Gender} shows the scores for models built with GloVe, ELMo and
BERT embeddings.  These numbers are roughly similar across models, and are far
from the desired values of $1$.  % , with only the Net Neutral score for ELMo
% reaching above $0.4$ to be significantly above random
This demonstrates gender bias in both static and contextualized embeddings.
Table~\ref{tbl:Ent-Gen} shows template fillers with the largest non-neutral
probabilities for GloVe.

\begin{table}[h!]
	\centering
	\resizebox{0.95\columnwidth}{!}{
	\begin{tabular}{ccccrr}
		%\hline
		\normalsize occ. & \normalsize verb & \normalsize obj. & \normalsize gen. & \normalsize ent. & \normalsize \hspace{-2mm}cont.\hspace{-1mm}
		\\ \hline
		banker & spoke to & crew & man & $0.98$ & \hspace{-2mm}$0.01$\hspace{-1mm} %\ENC{0.980}{0.018}{0.004}
		\\
		nurse & can afford & wagon & lady & $0.98$ & \hspace{-2mm}$0.00$\hspace{-1mm} %\ENC{0.980}{0.019}{0.001}
		\\
		librarian\hspace{-2mm} & spoke to & consul & woman & $0.98$  & \hspace{-2mm}0.00\hspace{-1mm} %\ENC{0.976}{0.022}{0.002}
		\\ \hline
		secretary & \hspace{-2mm}budgeted for & laptop & \hspace{-2mm}gentleman\hspace{-2mm} & $0.00$ & \hspace{-2mm}$0.99$\hspace{-1mm} 
		\\ 
		violinist & \hspace{-2mm}budgeted for & meal & \hspace{-2mm}gentleman\hspace{-2mm} & $0.00$ & \hspace{-2mm}$0.98$\hspace{-1mm} 
		\\ 
		\hspace{-1mm}mechanic & can afford & pig & lady & $0.00$ & \hspace{-2mm}$0.98$\hspace{-1mm}
		\\ \hline 
	\end{tabular}}
	\caption{\label{tbl:Ent-Gen}
		Gendered template parameters with largest entailment and contradiction values with the GloVe  model.  }
\end{table}

%%%%%%%%%%%%%%%%%%%%%%%%%%%%%%%%%%%%%%%%%%%%%%%%%%%%%%%%%%%%%%%%%%%%%
\subsection{Nationality and Religion}
\label{sec:rel+nat}

We can generate similar evaluations to measure bias related to
religions and nationalities.  Since the associated subspaces are not easily defined by term pairs, we 
%While there appears to be subspaces within
%word embeddings that capture both religions and nationalities (as
%discussed in a later section), they are not
%well-represented by term pairs (\eg, \example{man} and
%\example{woman}).  Rather we 
use a class of $32$ words
$\textsf{Demonyms}_{\textrm{Test}}$ (e.g., \textsf{French}) to
represent people from various nationalities.  Instead of comparing
these to occupations, we compare them to a term capturing polarity
(e.g., \textsf{evil}, \textsf{good}) from a \textsf{Polarity} set
with $26$ words, again in the extended version of the paper. 

Using the verb-object fillers as before (e.g., \textsf{crashed} a
\textsf{car}), we create sentence pairs such as
\begin{enumerate} [label=(\arabic*), leftmargin = 2.2em] \setcounter{enumi}{8}
	\item \label{ex:evil-car} The \textsf{evil} person \textsf{crashed} a \textsf{car}.
	\item The \textsf{French} person \textsf{crashed} a \textsf{car}.
\end{enumerate}
For a demonym $d \in \textsf{Demonym}_{\textrm{Test}}$, a
polarity term $p \in \textsf{Polarity}$, a verb 
$v \in \textsf{Verbs}$ and an object $o \in \textsf{Objects}$, we
generate a sentence pair as
\begin{enumerate} [label=(\arabic*), leftmargin = 2.2em] \setcounter{enumi}{10}
	\item \label{ex:p-a} The $p$ person $v$ a/an $o$.
	\item The $d$ person $v$ a/an $o$.
\end{enumerate}
and then 
%use one of the embedding-based models to 
generate the associated label probabilities, and compute the aggregate measures as before.

\begin{table}[h!]
	\centering
	\begin{tabular}{rcccc}
		%\hline
		Embedding & NN  &  FN & T:$0.5$ & T:$0.7$ 
		\\ \hline
		GloVe & 0.713 & 0.760 & 0.776 & 0.654 
		\\
		%GloVe(W) & 0.124 & 0.004 & 0.0 & 0.0 
		%\\
		ELMo & 0.698 & 0.776 & 0.757 & 0.597
		\\ \hline
	\end{tabular}
	\caption{\label{tbl:Nationality}
		Demonym-polarity neutrality scores, for models using GloVe and ELMo embeddings.}
\end{table}

Expanding all nationality templates provides
$26 \cdot 27 \cdot 95 \cdot 32 = 2{,}134{,}080$ entailment pairs.
Table \ref{tbl:Nationality} shows that for both GloVe and ELMo\footnote{
	Due to space constraints, for nationality and religion, we will focus on GloVe and
	ELMo embeddings. As we will see later, BERT presents technical challenges  for attenuating these biases.
},
the Net Neutral, Fraction Neutral, and Threshold (at 0.5 or 0.7) 
scores are between about $0.6$ and $0.8$.  While these scores are not
$1$, these do not numerically exhibit as much inherent bias as in
the gender case; the two tests are not strictly
comparable as the word sets are quite different.  Moreover, there
is still some apparent bias: for roughly $25\%$ of the sentence
pairs, something other than neutral was the most likely prediction.
The ones with largest non-neutral probabilities are shown in Table
\ref{tbl:Ent-Nat}.

\begin{table}[h!]
	\centering
	\resizebox{0.95\columnwidth}{!}{
	\begin{tabular}{ccccrr}
		%\hline
		\normalsize polar & \normalsize verb & \normalsize obj. & \normalsize dem. & \normalsize ent. & \normalsize \hspace{-2mm}cont.\hspace{-1mm}
		\\ \hline
		\hspace{-2mm}unprofessional\hspace{-3mm} & traded & brownie\hspace{-1mm} & \hspace{-3mm}Ukrainian\hspace{-4mm} & $0.97$ & \hspace{-3mm}$0.00$\hspace{-1mm} %\ENC{0.971}{0.025}{0.004}
		\\
		great & can afford\hspace{-2mm} & wagon & Qatari & $0.97$ & \hspace{-3mm}$0.00$\hspace{-1mm} %\ENC{0.970}{0.021}{0.008}
		\\
		\hspace{-2mm}professional & \hspace{-1mm}budgeted & auto & Qatari & $0.97$ & \hspace{-3mm}$0.01$\hspace{-1mm} %\ENC{0.972}{0.023}{0.006}
		\\ \hline
		evil & owns & oven & Canadian\hspace{-1mm} & $0.04$ & \hspace{-3mm}$0.95$\hspace{-1mm}
		\\
		evil & owns & phone & Canadian\hspace{-1mm} & $0.04$ & \hspace{-3mm}$0.94$\hspace{-1mm}
		\\
		smart & loved & urchin & Canadian\hspace{-1mm} & $0.07$ & \hspace{-3mm}$0.92$\hspace{-1mm}
		\\ \hline
	\end{tabular}}
	\caption{\label{tbl:Ent-Nat}
		Nationality template parameters with largest entailment and contradiction values with the GloVe model.  }
\end{table}

A similar set up is used to measure the bias associated
with Religions.  We use a word list of 17 adherents to religions \textsf{Adherent}$_\textrm{Test}$ such as \textsf{Catholic} to create sentences like
\begin{enumerate} [label=(\arabic*), leftmargin = 2.2em] \setcounter{enumi}{12}
	\item The \textsf{Catholic} person \textsf{crashed} a \textsf{car}.
\end{enumerate}
to be the paired hypothesis with sentence \ref{ex:evil-car}.  For each
adherent $h \in \textsf{Adherent}_{\textrm{Test}}$, a polarity term
$p \in \textsf{Polarity}$, verb $v \in \textsf{Verbs}$ and object $o
\in \textsf{Objects}$, we generate a sentence pair in the form of sentence \ref{ex:p-a} and
\begin{enumerate} [label=(\arabic*), leftmargin = 2.2em] \setcounter{enumi}{13}
	\item The $h$ person $v$ a/an $o$.
\end{enumerate}
We aggregated the predictions under our measures as before. 
Expanding all religious templates provides
$26 \cdot 27 \cdot 95 \cdot 17 = 1{,}133{,}730$ 
entailment pairs.  
The results for GloVe- and ELMo-based inference are shown in Table \ref{tbl:Religion}.  We observe a similar pattern as with Nationality, with about $25\%$ of the sentence pairs being inferred as non-neutral; the largest non-neutral template expansions are in Table \ref{tbl:Ent-Rel}.   The biggest difference is that the ELMo-based model performs notably worse on this test.  

\begin{table}[h!]
	\centering
	\begin{tabular}{rcccc}
		%\hline
		Embedding & NN  &  FN & T:$0.5$ & T:$0.7$ 
		\\ \hline
		GloVe & 0.710 & 0.765 & 0.785 & 0.636 
		\\
		ELMo & 0.635 & 0.651 & 0.700 & 0.524
		\\ \hline
	\end{tabular}
	\caption{\label{tbl:Religion}
		Religion-polarity neutrality scores, for models using GloVe and ELMo embeddings.}
\end{table}

%\vs{This would be a good place to put some examples with high entail or contradict probabilities for religions}

\begin{table}[h!]
	\centering
	\resizebox{0.95\columnwidth}{!}{
	\begin{tabular}{ccccrr}
		%\hline
		\normalsize polar & \normalsize verb & \normalsize obj. & \normalsize adh. & \normalsize ent. & \normalsize \hspace{-1mm}cont.\hspace{-1mm}
		\\ \hline
		\hspace{-1mm}dishonest & sold & calf & satanist & $0.98$ & \hspace{-1mm}0.01\hspace{-1mm} %\ENC{0.976}{0.019}{0.005}
		\\
		\hspace{-1mm}dishonest & \hspace{-2mm}swapped\hspace{-2mm} & cap & Muslim & $0.97$ & \hspace{-1mm}0.01\hspace{-1mm} %\ENC{0.972}{0.023}{0.005}
		\\
		ignorant & hated & owner & Muslim & $0.97$ & \hspace{-1mm}0.00\hspace{-1mm} %\ENC{0.971}{0.025}{0.004}
		\\ \hline
		smart & saved & dresser & Sunni & $0.01$  & \hspace{-1mm}0.98\hspace{-1mm}
		\\
		\hspace{-1mm}humorless & saved & potato & Rastafarian & $0.02$  & \hspace{-1mm}0.97\hspace{-1mm}
		\\
		terrible & saved & lunch & \hspace{-2mm}Scientologist\hspace{-2mm} & $0.00$  & \hspace{-1mm}0.97\hspace{-1mm}
		\\ \hline
	\end{tabular}}
	\caption{\label{tbl:Ent-Rel}
		Religion template parameters with largest entailment and contradiction values with the GloVe model.  }
\end{table}

\section{Attenuating Bias in Static Embeddings}
\label{sec:Remove-GloVe}

We saw above that several kinds of biases exist in
static embeddings (specifically GloVe). We can to some extent
attenuate it.  For the case of gender, this comports with the
effectiveness of debiasing on previously studied intrinsic measures
of bias~\cite{debias,Bias1}.  We focus on the simple
\emph{projection} operator~\cite{Bias1} which simply identifies a
subspace associated with a concept hypothesized to carry bias, and
then removes that subspace from \emph{all} word representations.
Not only is this approach simple and outperforms other approaches on
intrinsic measures~\cite{Bias1}, it also does not have the potential
to leave residual information among associated
words~\cite{gonen2019lipstick} unlike hard debiasing~\cite{debias}.
% Moreover, it does not require a fixed subset of words to
% operate on, and thus has the potential to generalize to contextual
% embeddings.  
There are also retraining-based
mechanisms~\cite[e.g.]{gn-glove}, but given that building
word embeddings can be prohibitively expensive, we focus on the much
simpler post-hoc modifications.

%%%%%%%%%%%%%%%%%%%%%%%%%%%%%%%%%%%%%%%%%%%%%%%%%%%%%%%%%%%%%%%%%%%%%
\subsection{Bias Subspace}
\label{sec:G-bias-dir}

For the gender direction, we identify a bias subspace using only the embedding of the words \texttt{he} and \texttt{she}.  This provides a single bias vector, and is a strong single direction correlated with other explicitly gendered words.  Its cosine similarity with the two-means vector from \textsf{Names} used in~\cite{Bias1} is $0.80$ and with \textsf{Gendered} word pairs from~\cite{debias}  is $0.76$.  
%Moreover, the second principal value among gendered words is $0.57$ times the first one, a significant drop.  
%, indicating that it is a fairly significant 1-dimensional subspace, and observed in these previous papers to be effective to attenuate bias.   

For nationality and religion, the associated directions are present and have similar traits to the gendered one (Table \ref{tbl:G-stable}), but are not quite as simple to work with.
For nationalities, we identify a separate set of $8$ demonyms than those used to create sentence pairs as \textsf{Demonym}$_{\textrm{Train}}$, and use their first principal component to define a $1$-dimensional demonym subspace.  
For religions, we similarly use a \textsf{Adherent}$_{\textrm{Train}}$ set, again of size $8$, but use the first $2$ principal components to define a $2$-dimensional religion subspace.  
In both cases, these were randomly divided from full sets \textsf{Demonym} and \textsf{Adherent}. %(the union of Test and Train subsets).  
%These subspace as far less stable than the ones identifying gender.  
Also, the cosine similarity of the top singular vector from the full sets with that derived from the training set was $0.56$ and $0.72$ for demonyms and adherents, respectively.  Again, there is a clear correlation, but perhaps slightly less definitive than gender.  
%Table \ref{tbl:G-stable} shows the percentages of top singular vectors derived from the full sets.  

\begin{table}[h!]
	\centering
	\begin{tabular}{rcccr}
		%\hline
		Embedding & 2nd  &  3rd & 4th & cosine 
		\\ \hline
		\textsf{Gendered} & 0.57 & 0.39 & 0.24 & 0.76
		\\
		\textsf{Demonyms} & 0.45 & 0.39 & 0.30 & 0.56 
		\\
		\textsf{Adherents} & 0.71 & 0.59 & 0.4 & 0.72
		\\ \hline
	\end{tabular}
	\caption{\label{tbl:G-stable}
		Fraction of the top principal value with the $x$th principal value with the GloVe embedding for \textsf{Gendered}, \textsf{Demonym}, and \textsf{Adherent} datasets.  The last column is the cosine similarity of the top principal component with the derived subspace.  }
\end{table}

%%%%%%%%%%%%%%%%%%%%%%%%%%%%%%%%%%%%%%%%%%%%%%%%%%%%%%%%%%%%%%%%%%%%%
\subsection{Results of Bias Projection}
\label{sec:G-project}

By removing these derived subspaces from GloVe, we demonstrate significant decrease in bias.
Let us start with gender, where we removed  the \textsf{he}-\textsf{she} direction, and then recomputed the various bias scores.  Table \ref{tbl:G-Gen-remove} shows these results, as well as the effect of projecting a random vector (averaged over 8 such vectors), along with the percent change from the original GloVe scores. We see that the scores increase between $25\%$ and $160\%$ which is quite significant compared to the effect of random vectors which range from decreasing $6\%$ to increasing by $3.5\%$.  

\begin{table}[h!]
	\centering
	\begin{tabular}{rcccc}
		\multicolumn{5}{c}{Gender (GloVe)} \\
		%\hline
		& NN  &  FN & T:$0.5$ & T:$0.7$ 
		\\ \hline
		proj & 0.480 & 0.519 & 0.474 & 0.297 
		\\
		diff & +24.7\% & +31.7\% & +41.9\% & +160.5\%
		\\ \hline
		rand & 0.362 & 0.405 & 0.323 & 0.118 
		\\
		diff & -6.0\% & +2.8\% & -0.3\% & +3.5\%
		\\ \hline
	\end{tabular}
	\caption{\label{tbl:G-Gen-remove}
		Effect of attenuating gender bias using the \textsf{he}-\textsf{she} vector, and random vectors with difference (diff) from no attenuation.}
\end{table}

For the learned demonym subspace, the effects are shown in Table \ref{tbl:G-DemRel-remove}.  Again, all the neutrality measures are increased, but more mildly.  The percentage increases range from $13$ to $20\%$, but this is expected since the starting values were already larger, at about $75\%$-neutral; they are now closer to $80$ to $90\%$ neutral.

\begin{table}[h!]
	\centering
	\begin{tabular}{rcccc}
		\multicolumn{5}{c}{Nationality (GloVe)} \\
		%\hline
		& NN  &  FN & T:$0.5$ & T:$0.7$ 
		\\ \hline
		proj & 0.808 & 0.887 & 0.910 & 0.784 
		\\
		diff & +13.3\% & +16.7\% & +17.3\% & +19.9\%
		\\ \hline
		\multicolumn{5}{c}{Religion (GloVe)} \\
		%\hline
		& NN  &  FN & T:$0.5$ & T:$0.7$ 
		\\ \hline
		proj & 0.794& 0.894 & 0.913 & 0.771 
		\\
		diff & +11.8\% & +16.8\% & +16.3\% & +21.2\%
		\\ \hline
	\end{tabular}
	\caption{\label{tbl:G-DemRel-remove}
		Effect of attenuating nationality bias using the 
		\textsf{Demonym}$_{\textrm{Train}}$-derived vector, and religious bias using the 
		\textsf{Adherent}$_{\textrm{Train}}$-derived vector, with difference (diff) from no attenuation.}
\end{table}

The results after removing the learned adherent subspace, as shown in Table \ref{tbl:G-DemRel-remove} are quite similar as with demonyms.  The resulting neutrality scores and percentages are all similarly improved, and about the same as with nationalities.

%\begin{table}[h!]
%	\centering
%	\begin{tabular}{rcccc}
%		\multicolumn{5}{c}{Religion (GloVe)} \\
%		%\hline
%		& NN  &  FN & T:$0.5$ & T:$0.7$ 
%		\\ \hline
%		proj & 0.794& 0.894 & 0.913 & 0.771 
%		\\
%		diff & +11.8\% & +16.8\% & +16.3\% & +21.2\%
%		\\ \hline
%	\end{tabular}
%	\caption{\label{tbl:G-Rel-remove}
%		Effect of attenuating religious bias using the 
%		\textsf{Adherent}$_{\textrm{Train}}$-derived vector with difference (diff) from no attenuation.}
%\end{table}

Moreover, the dev and test scores (Table \ref{tbl:G-dev+test}) on the SNLI benchmark is $87.81$ and $86.98$ before, and $88.14$ and $87.20$ after the gender projection.  So the scores actually improve slightly after this bias attenuation!  
For the demonyms and religion, the dev and test scores show very little change.  

\begin{table}[h]
	\centering
	\begin{tabular}{rcccc}
		\multicolumn{5}{c}{SNLI Accuracies (GloVe)} \\
		%\hline 
		& orig & -gen & -nat & -rel
		\\ \hline
		Dev & 87.81 & 88.14 & 87.76 & 87.95
		\\
		Test & 86.98  & 87.20 & 86.87 &  87.18
		\\ \hline
	\end{tabular}
	\caption{\label{tbl:G-dev+test}
		SNLI dev/test accuracies before debiasing GloVe embeddings (orig) and after debiasing gender, nationality, and religion.}
\end{table}

%%%%%%%%%%%%%%%%%%%%%%%%%%%%%%%%%%%%%%%%%%%%%%%%%%%%%%%%%%%%%%%%%%%%%
%%%%%%%%%%%%%%%%%%%%%%%%%%%%%%%%%%%%%%%%%%%%%%%%%%%%%%%%%%%%%%%%%%%%%
\section{Attenuating Bias in Contextualized Embeddings}
\label{sec:Remove-ELMo}

%In this section, with less success, we attempt to attenuate bias in contextualized word vector embeddings (specifically ELMo and BERT).

Unlike GloVe, ELMo and BERT are context-aware dynamic embeddings that are computed using multi-layer encoder modules over the sentence.
For ELMo this results in three layers of embeddings, each $1024$-dimensional.
The first layer---a character-based
model---is essentially a static word embedding and all three are
interpolated as word representations for the NLI model.
Similarity, BERT (the base version) has $12$-layer contextualized embeddings, each $768$-dimensional.
Its input embeddings are also static.
% Before inference tasks are performed, a linear combination of
% these layers is learned to create a single $1024$-dimensional
% embedding, where typically layer $2$ has the largest weight.
%
We first investigate how to address these issues on ELMo, and then extend it to BERT which has the additional challenge that the base layer only embeds representations for subwords.  

%%%%%%%%%%%%%%%%%%%%%%%%%%%%%%%%%%%%%%%%%%%%%%%%%%%%%%%%%%%%%%%%%%%%%
\subsection{ELMo All Layer Projection: Gender}
\label{sec:all-layer}
Our first attempt at attenuating bias is by directly replicating the projection procedure where we learn a bias subspace, and remove it from the embedding.  The first challenge is that each time a word appears, the context is different, and thus its embedding in each layer of a contextualized embedding is different.  

However, we can embed the 1M sentences in a representative training corpus WikiSplit\footnote{\url{https://github.com/google-research-datasets/wiki-split}}, and average  embeddings of word types.  This averages out contextual information and incorrectly blends senses; but this process does not re-position these words. This process can be used to learn a subspace, say encoding gender and is successful at this task by intrinsic measures: on ELMo the second singular value of the full \textsf{Gendered} set is $0.46$ for layer 1, $0.36$ for layer 2, and $0.47$ for layer 3, all sharp drops.

Once this subspace is identified, we can then apply the projection operation onto each layer individually.  Even though the embedding is contextual, this operation makes sense since it is applied to all words; it just modifies the ELMo embedding of any word (even ones unseen before or in new context) by first applying the original ELMo mechanism, and then projecting afterwards.  

However, this does not significantly change the neutrality on gender specific inference task.  Compared to the original results in Table \ref{tbl:Gender} the change, as shown in Table \ref{tbl:E-Gen-all-remove} is not more, and often less than, projecting along a random direction (averaged over 4 random directions).  
We conclude that despite the easy-to-define gender direction, this mechanism is not effective in attenuating bias as defined by NLI tasks.  
We hypothesize that the random directions work surprisingly well because it destroys some inherent structure in the ELMo process, and the prediction reverts to neutral.

\begin{table}[h!]
	\centering
	\begin{tabular}{rcccc}
		\multicolumn{5}{c}{Gender (ELMo All Layers)} \\
		%\hline
		& NN  &  FN & T:$0.5$ & T:$0.7$ 
		\\ \hline
		proj & 0.423 & 0.419 & 0.363 & 0.079 
		\\
		diff & +1.6\% & + 7.2\% & + 19.8\% & + 25.4\%
		\\ \hline
		rand & 0.428 & 0.412 & 0.372 & 0.115 
		\\
		diff & +2.9\% & +5.4\% & +22.8\% & +82.5\%
		\\ \hline
	\end{tabular}
	\caption{\label{tbl:E-Gen-all-remove}
		Effect of attenuating gender bias on \emph{all layers} of ELMo and with random vectors with difference (diff) from no attenuation.}
\end{table}

%%%%%%%%%%%%%%%%%%%%%%%%%%%%%%%%%%%%%%%%%%%%%%%%%%%%%%%%%%%%%%%%%%%%%
\subsection{ELMo Layer 1 Projection: Gender}
\label{sec:layer1-Gen}

Next, we show how to significantly attenuate gender bias in ELMo embeddings: we invoke the projection mechanism, but only on layer 1. The layer is a static embedding of each word -- essentially a look-up table for words independent of context.  Thus, as with GloVe we can find a strong subspace for gender using only the \textsf{he-she} vector.  Table \ref{tbl:E-stable} shows the stability of the subspaces on the ELMo layer 1 embedding for \textsf{Gendered} and also \textsf{Demonyms} and \textsf{Adherents}; note this fairly closely matches the table for GloVe, with some minor trade-offs between decay and cosine values.   
%For the gender subspace the singular value decay is more rapid than in GloVe, but the cosine similarity is not as large.  
%The demonym subspaces does not have singular values decay as rapidly, but has a larger cosine similarity with the split compared to GloVe; and the adherent subspace has a lower cosine similarity, but more-rapid singular value decay, compared to GloVe.  

\begin{table}[h!]
	\centering
	\begin{tabular}{rcccr}
		%\hline
		Embedding & 2nd  &  3rd & 4th & cosine 
		\\ \hline
		\textsf{Gendered} & 0.46 & 0.32 & 0.29 & 0.60
		\\
		\textsf{Demonyms} & 0.72 & 0.61 & 0.59 & 0.67 
		\\
		\textsf{Adherents} & 0.63 & 0.61 & 0.58& 0.41
		\\ \hline
	\end{tabular}
	\caption{\label{tbl:E-stable}
		Fraction of the top principal value with the $x$th principal value with the ELMo layer 1 embedding for  \textsf{Gendered}, \textsf{Demonym}, and \textsf{Adherent} datasets.  The last column shows the cosine similarity of the top principal component with the derived subspace.  }
\end{table}

Once this subspace is identified, we apply the projection operation on the resulting layer 1 of ELMo.  We do this before the BiLSTMs in ELMo generates the layers 2 and 3.
The resulting full ELMo embedding attenuates intrinsic bias at layer 1, and then generates the remainder of the representation based on the learned contextual information.  We find that perhaps surprisingly when applied to the gender specific inference tasks, that this indeed increases neutrality in the predictions, and hence attenuates bias.  

Table \ref{tbl:E-Gen-1-remove} shows that each measure of neutrality is significantly increased by this operation, whereas the projection on a random vector (averaged over 8 trials) is within $3\%$ change, some negative, some positive.  For instance, the probability of predicting neutral is now over $0.5$, an increase of $+28.4\%$, and the fraction of examples with neutral probability  $>0.7$ increased from $0.063$ (in Table \ref{tbl:Gender}) to $0.364$ (nearly a $500\%$ increase).

\begin{table}[h!]
	\centering
	\begin{tabular}{rcccc}
		\multicolumn{5}{c}{Gender (ELMo Layer 1)} \\
		%\hline
		& NN  &  FN & T:$0.5$ & T:$0.7$ 
		\\ \hline
		proj & 0.488 & 0.502 & 0.479 & 0.364 
		\\
		diff & +17.3\% & +28.4\% & +58.1\% & +477.8\%
		\\ \hline
		rand & 0.414 & 0.402 & 0.309 & 0.062 
		\\
		diff & -0.5\% & +2.8\% & +2.0\% & -2.6\%
		\\ \hline
	\end{tabular}
	\caption{\label{tbl:E-Gen-1-remove}
		Effect of attenuating gender bias on \emph{layer 1} of ELMo with \textsf{he-she} vectors and random vectors with difference (diff) from no attenuation.}
\end{table}

%%%%%%%%%%%%%%%%%%%%%%%%%%%%%%%%%%%%%%%%%%%%%%%%%%%%%%%%%%%%%%%%%%%%%
\subsection{ELMo Layer 1 Projection: Nationality \& Religion}
\label{sec:layer1-Rel+Nat}

We next attempt to apply the same mechanism (projection on layer 1 of ELMo) to the subspaces associated with nationality and religions, but we find that this is not effective.  
%We use nationality and religion vectors defined the same way as with GloVe (on a training set of demonyms and adherents), and with stability statistics shown in Table \ref{tbl:E-stable}.  And then apply the projection operation on layer 1 of ELMo before using this to create the layer 2 and 3 embeddings.  

The results of the aggregate neutrality of the nationality and religion specific inference tasks are shown in Table \ref{tbl:E-NatRel-1-remove}, respectively.  The neutrality actually decreases when this mechanism is used.  This negative result indicates that simply reducing the nationality or religion information from the first layer of ELMo does not help in attenuating the associated bias on inference tasks on the resulting full model.  

\begin{table}[h!]
	\centering
	\begin{tabular}{rcccc}
		\multicolumn{5}{c}{Nationality (ELMo Layer 1)} \\
		%\hline
		& NN  &  FN & T:$0.5$ & T:$0.7$ 
		\\ \hline
		proj & 0.624 & 0.745 & 0.697 & 0.484 
		\\
		diff & -10.7\% & -4.0\% & -7.9\% & -18.9\%
		\\ \hline
		\multicolumn{5}{c}{Religion (ELMo Layer 1)} \\
		%\hline
		& NN  &  FN & T:$0.5$ & T:$0.7$ 
		\\ \hline
		proj & 0.551 & 0.572 & 0.590 & 0.391 
		\\
		diff & -13.2\% & -12.1\% & -15.7\% & -25.4\%
		\\ \hline
	\end{tabular}
	\caption{\label{tbl:E-NatRel-1-remove}
		Effect of attenuating nationality bias on \emph{layer 1} of ELMo with the demonym direction, and religious bias with the adherents direction, with difference (diff) from no attenuation.}
\end{table}

%\begin{table}[h!]
%	\centering
%	\begin{tabular}{rcccc}
%		\multicolumn{5}{c}{Religion (ELMo Layer 1)} \\
%		\hline
%		& NN  &  FN & T:$0.5$ & T:$0.7$ 
%		\\ \hline
%		proj & 0.551 & 0.572 & 0.590 & 0.391 
%		\\
%		diff & -13.2\% & -12.1\% & -15.7\% & -25.4\%
%		\\ \hline
%	\end{tabular}
%	\caption{\label{tbl:E-Rel-1-remove}
%		Effect of attenuating religion bias on \emph{layer 1} of ELMo with the adherents direction with difference (diff) from no %attenuation.}
%\end{table}

We have several hypotheses for why this does not work.  Since these scores have a higher starting point than on gender, this may distort some information in the ultimate ELMo embedding, and the results are reverting to the mean.  
Alternatively, layers 2 and 3 of ELMo may be (re-)introducing bias into the final word representations from the context, and this effect is more pronounced for nationality and religions than gender.

We also considered that the learned demonym or adherent subspace on the training set is not good enough to invoke the projection operation as compared to the gender variant.  However, we tested a variety of other ways to define this subspace, including using country and religion names (as opposed to demonyms and adherents) to learn the nationality and religion subspaces, respectively. This method is supported by the linear relationships between analogies encoded shown by static word embeddings~\cite{Mik1}.  
While in a subset of measures this did slightly better than using a separate training and test set for just the demonyms and adherents, it does not have more neutrality than the original embedding.  
Even training the subspace \emph{and} evaluating on the full set of \textsf{Demonyms} and \textsf{Adherents} does not increase the measured aggregate neutrality scores.

%\begin{table*}[]
%     \begin{centering}
%	\begin{tabular}{rcccccc}
%		\hline
%		& NN  &  FN & T:$0.5$ & T:$0.7$  & dev & test
%		\\ \hline
%		BERT & 0.421 & 0.397 & 0.374 & 0.209 &0.903 & 0.9022
%		\\
%		\hline
%		(1) BERT +FT + debiased& 0.396 & 0.371 & 0.341 & 0.167 & 0.903 & 0.9017
%		\\
%		\% & -5.9\% & -6.5\% & -8.8\% & -20\% & 0\% & -0.05\% \\
%			BERT + FT rand debiased &0.398 & 0.388 &0.328 & 0.201 & 0.9044 & 0.898 \\
%			\% & -5.4\% &-2.3\% & -12.3\% & -3.8\% & +0.15\% & -0.4\%\\ 
%				\hline
%
%			(2) BERT debiased + FT + debiased  & 0.516 & 0.526 & 0.501 & 0.354 & 0.907 & 0.9023\\
%			\% & +22.6\% & +32.4\% & +33.9\% & +69.4\% & +0.4\% & +0.01\% \\
%		BERT rand + FT + rand debiased & 0.338 & 0.296  & 0.253 & 0.168 & 0.905 & 0.9013\\
%		\% & -19.7\% & -25.4\% & -32.6\% & -19.6\% & +0.2\% & -0.1\%
%		\\ \hline
%
%		(3) BERT  debiased + FT & 0.521 & 0.537  & 0.520 & 0.355 & 0.903 & 0.9015
%		 \\
%		 \% & +23.7\% & +35.2\% & +39\% & +69.8\% & 0\% & -0.07\% \\
%		 BERT rand + FT & 0.335 & 0.299  & 0.221 & 0.159 & 0.903 & 0.9013
%		 \\ 
%		 \% & -20.4\% & -24.7\% & -40.9\% & -23.9\% & 0\% & -0.1\%
%		\\ \hline
%	\end{tabular}
%	\caption{\label{tbl:bert} BERT scores}
%	\end{centering}
%\end{table*}

\begin{table}[]
	\centering
	\resizebox{0.95\columnwidth}{!}{
	\begin{tabular}{rcccc}
		\multicolumn{5}{c}{Gender (BERT)} \\
		%\hline
		& NN  &  FN & T:$0.5$ & T:$0.7$  
		\\ \hline
		no proj & 0.421 & 0.397 & 0.374 & 0.209 
		\\
		\hline
		proj@test & 0.396 & 0.371 & 0.341 & 0.167 
		\\
		diff & -5.9\% & -6.5\% & -8.8\% & -20\% 
		\\
		rand@test &0.398 & 0.388 &0.328 & 0.201 
		\\		
		diff & -5.4\% &-2.3\% & -12.3\% & -3.8\% 
		\\\hline
		proj@train/test  & 0.516 & 0.526 & 0.501 & 0.354 
		\\
		diff & +22.6\% & +32.4\% & +33.9\% & +69.4\% 
		\\
		rand & 0.338 & 0.296  & 0.253 & 0.168 
		\\
		diff & -19.7\% & -25.4\% & -32.6\% & -19.6\% 
		\\ \hline
	\end{tabular}}
	\caption{\label{tbl:bert} 		
		The effect of attenuating gender bias on \emph{subword embeddings} in BERT with the \textsf{he}-\textsf{she} direction and random vectors with difference (diff) from no attenuation.}
\end{table}

\subsection{BERT Subword Projection}
We next extend the debiasing insights learned on ELMo and apply them to BERT~\cite{bert}. In addition to being contextualized, BERT presents two challenges for debiasing. First, unlike ELMo, BERT operates upon subwords (e.g, \textsf{ko}, \textsf{--sov}, and \textsf{--ar} instead of the word \textsf{Kosovar}).  This makes identifying the subspace associated with nationality and religion even more challenging, and thus we leave addressing this issue for future work.  However, for gender, the simple pair \textsf{he} and \textsf{she} are each subwords, and can be used to identify a gender subspace in the embedding layer of BERT, and this is the only layer we apply the projection operation. Following the results from ELMo, we focus on debiasing the context-independent subword BERT embeddings by projecting them along a pre-determined gender direction. 

A second challenge concerns {\em when} the debiasing step should be applied. Pre-trained BERT embeddings are typically treated as an initialization for a subsequent fine-tuning step that adapts the learned representations to a downstream task (e.g., NLI). We can think of the debiasing projection as a constraint that restricts what information from the subwords is available to the inner layers of BERT.
Seen this way, two options naturally present themselves for when the debiasing operation is to be performed. We can either (1) fine-tune the NLI model without debiasing and impose the debiasing constraint {\em only} at test time, or,
(2) apply debiasing both when the model is fine-tuned, and also at test time.

% \jeff{VIVEK:  Rewrite so (3) is not natural -- we now omit}
% A second challenge with BERT is that the pre-trained embeddings are not recommended for use ``out of the box'', instead a fine-tuning step on the downstream task (\eg NLI) is used to achieve state-of-the-art results. So we need to consider how this interacts with the fine-tuning step.  
% We consider the following options: 
%  (1) fine-tune first, and during evaluation, project along the gender direction computed from pre-trained BERT.  
%  (2) fine-tune with gender projection computed from pre-trained BERT, then apply the same projection during evaluation.
%  (3) fine-tune with gender projection computed from pre-trained BERT, then during evaluation, apply no gender projection.
%  (4) fine-tune with gender projection that is dynamically computed from the running BERT, then during evaluation, apply the gender projection computed from the fine-tuned BERT.

Our evaluation, shown in Table \ref{tbl:bert}, show that method (1) (debias@test) is ineffective at debiasing with gender as measured using NLI; however, that method (2) (debias@train/test) is effective at reducing bias.  In each case we compare against projecting along a random direction (repeated 8 times) in place of the gender direction (from he-she).  These each have negligible effect, so these improvements are significant.  Indeed the results from method (2) result in the least measured NLI bias among all methods while retaining  test scores on par with the baseline BERT model.

\section{Discussions, Related works \& Next Steps}
\label{sec:discussion}
% In this section, we will situate our work in the broader context of
% research in debiasing NLP, leading to several open questions and
% possible research directions.

\paragraph{Glove vs. ELMo vs. BERT}

While the mechanisms for attenuating bias of ELMo
were not universally successful, they were always successful on GloVe.  
Moreover, the overall neutrality scores are higher on (almost) all tasks on the debiased GloVe embeddings than ELMo.
Yet, GloVe-based models underperform ELMo-based models on NLI test
scores.  

Table~\ref{tbl:E-dev+test} summarizes the dev and test scores for ELMo.
%
% The baseline dev and test scores for the ELMo-based model are $89.03$ and $88.37$, respectively.  
% After debiasing for gender on layer 1, the scores drop fairly insignificantly to $88.77$ and $88.04$; but after debiasing on all layers they drop significantly to $88.36$ and $87.42$, indicating that the projection operation on all layers is causing a more significant alteration to the ELMo embeddings. 
% Similarly for debiasing with respect to nationality on layer 1 the scores drop to $89.01$ and $87.99$;
% and for debiasing  for religions on layer 1 the scores drop to $89.04$ and $88.30$.  
We see that the effect of debiasing 
is fairly minor on the original prediction goal, and these scores remain slightly larger
than the models based on GloVe, both before and after
debiasing. These observations suggest that while ELMo offers better
predictive accuracy, it is also harder to debias
than simple static embeddings.

%dev 88.36, test 87.42

\begin{table}[h]
	\centering
	\begin{tabular}{rccccc}
		\multicolumn{6}{c}{SNLI Accuracies (ELMo)} \\
		%\hline 
		& orig & \hspace{-1mm}-gen(all)\hspace{-1mm} & -gen(1)\hspace{-1mm} & -nat(1)\hspace{-1mm} & -rel(1)\hspace{-1mm} 
		\\ \hline
		\hspace{-1mm}Dev & 89.03& $88.36$ & 88.77 & 89.01 & 89.04\hspace{-1mm} 
		\\
		\hspace{-1mm}Test & 88.37 & $87.42$ & 88.04 & 87.99 & 88.30\hspace{-1mm} 
		\\ \hline
	\end{tabular}
	\caption{\label{tbl:E-dev+test}
		SNLI dev/test accuracies before debiasing ELMo (orig) and after debiasing gender on all layers and layer 1, debiasing nationality and religions on layer 1.}
\end{table}

Overall, on gender, however, BERT provides the best dev and test scores ($90.70$ and $90.23$) while also achieving the highest neutrality scores, see in Table \ref{tbl:gen-best-scores}.  Recall we did not consider nationalities and religions with BERT because we lack a method to define associated subspaces to project.  

\begin{table}[h]
	\centering
	\resizebox{0.95\columnwidth}{!}{
	\begin{tabular}{rcccccc}
		\multicolumn{7}{c}{Gender (Best)} \\
		%\hline 
		& NN  &  FN & T:$0.5$ & T:$0.7$ & dev & test
		\\ \hline
		GloVe & 0.480 & 0.519 & 0.474 & 0.297 & 88.14 & 87.20 
		\\
		ELMo & 0.488 & 0.502 & 0.479 & 0.364 & 88.77 & 88.04 
		\\
		BERT & 0.516 & 0.526 & 0.501 & 0.354 & 90.70 & 90.23
		\\ \hline
	\end{tabular}}
	\caption{\label{tbl:gen-best-scores}
		Best effects of attenuating gender bias for each embedding type.  The dev and test scores for BERT before debiasing are $90.30$ and $90.22$ respectively.} %The change in scores due to debiasing is negligible.  }
\end{table}

\paragraph{Further resolution of models and examples.}
Beyond simply measuring the error in aggregate over all templates,
and listing individual examples, there are various interesting
intermediate resolutions of bias that can be measured.  We can, for
instance, restrict to all nationality templates which involve
\textsf{rude} $\in$ \textsf{Polarity} and \textsf{Iraqi} $\in$
\textsf{Demonym}, and measure their average entailment: in the GloVe
model it starts as $99.3$ average entailment, and drops to $62.9$
entailment after the projection of the demonym subspace.  
%Similarly,
%in the religion templates the average entailment associated with
%\example{stupid} $\in$ \textsf{Polarity} and \example{satanist}
%$\in$ \textsf{Adherent} in the GloVe model drops from $76.8$ average
%entailment to $0.002$ average entailment after the projection of the
%adherent subspace.

% If it makes an error, it should make the same error for all elements.  E.g., stupid lybian may be an error.  But if its not also the same error for stupid canadian (and many others), it implies that there may be more specific 

%In addition to evaluating all templates in aggregate (as proposed and reported on broadly above) and individual sentence pairs, it is also possible to consider individual word pairs  (e.g., \example{stupid} and \example{libyan}) and repeat out aggregate experiments over all templates which vary these words (i.e., over all \textsf{Verbs} and \textsf{Objects}).  This provides a specific sets of bias over pairs of words.  

% Next steps:  This just shows the non-neutrality of words.  But it does not show the consistency.  Perhaps a \example{lybian} could be strongly entailed with both \example{good} and \example{bad}.  

\paragraph{Sources of bias.} Our bias probes run the risk of
entangling two sources of bias: from the representation, and from
the data used to train the NLI
task. \cite{rudinger2017social},~\cite{gururangan2018annotation} and
references therein point out that the mechanism for gathering the
SNLI data allows various stereotypes (gender, age, race, etc.) and
annotation artifacts to seep into the data. What is the source of  the non-neutral inferences? % The construction of the
% entailment pairs using templates keeps all other aspects of a
% sentence constant except for the probe words that are placed in the
% subject position. The fact that the pair
% \example{rude}-\example{iraqi} shows a much higher probability for
% entailment than other nationalities suggests that the problem is in
% the way the words are encoded. Furthermore, 
The observation from GloVe that the
three bias measures can increase by attenuation strategies that {\em
only} transform the word embeddings indicates that any bias that
may have been removed is from the word embeddings.
%
%Of course, after the attenuation operations, the resulting models are still imperfect (especially ELMo). 
The residual bias could
still be due to word embeddings, or as the literature points out,
from the SNLI data. Removing the latter is an open question; we
conjecture that it may be possible to design loss functions that
capture the spirit of our evaluations in order to address such bias.

\paragraph{Relation to error in models.}
A related concern is that the examples of non-neutrality observed in
our measures are simply model errors.  We argue this
is not so for several reasons.  First, the probability of
predicting neutral is below (and in the case of gendered examples,
far below $40 - 50\%$) the scores on the test sets (almost
$90\%$), indicating that these examples pose problems beyond the
normal error.  Also, through the projection of random directions
in the embedding models, we are essentially measuring a type of random
perturbations to the models
themselves; the result of this perturbation is fairly insignificant,
indicating that these effects are real.

\paragraph{Biases as invalid inferences.} 
We use NLI to
measure bias in word embeddings. The definition of the NLI task
lends itself naturally to identifying biases. Indeed, the ease with
which we can reduce other reasoning tasks to textual entailment was
a key motivation for the various PASCAL entailment
challenges \cite[\emph{inter alia}]{dagan2006the-pascal}. While, we have explored three kinds of biases that have important
societal impacts, the mechanism is easily extensible to other
types of biases.  

\paragraph{Relation to coreference resolution as a measure of bias.}
Coreference resolution, especially pronoun coreference, has been recently used
as an extrinsic probe to measure bias in representations~\cite[for example]{rudinger2018gender,zhao2019gender,gap}. This direction is complementary to our work; making an incorrect coreference decision constitutes an invalid inference. However, coreference resolution may be a difficult probe to realize because the task itself is considered to be an uphill challenge in NLP. Yet, we believe that these two tasks can supplement each other to provide a more robust evaluation metric for bias.

% While it is another downstream task complementary to textual entailment in measurement of bias, it is also a much harder probe which makes it more prone to noise. % Further, it is not clear if along with named entities, entities like \example{doctor} or other occupation terms can also be coreference resolved.

% \paragraph{Attenuation vs. Retraining.} 

\section{Conclusion}
\label{sec:conclusion}

In this paper, we use the observation that biased representations
lead to biased inferences to construct a systematic probe for
measuring biases in word representations using the task of natural
language inference. Our experiments using this probe reveal that
GloVe, ELMo, and BERT embeddings all encode gender, religion and
nationality biases.  We explore the use of a projection-based method
for attenuating biases. Our experiments show that the method works for
the static GloVe embeddings. We extend the approach to
contextualized embeddings (ELMo, BERT) by debiasing the first (non-contextual)
layer alone and show that for the well-characterized gender direction, this
simple approach can effectively attenuate bias in both contextualized embeddings
without loss of entailment accuracy.

\section{Acknowledgments}
Thanks to NSF CCF-1350888, ACI-1443046, CNS-1514520, CNS-1564287, and  IIS-1816149, and  SaTC-1801446, Cyberlearning-1822877 and a generous gift from Google.

\bibliographystyle{plain}
\bibliography{icpbib}

\newpage

\large{\textbf{On Measuring and Mitigating Biased Inferences of Word Embeddings -- Supplementary Material}}

\section{Word Lists}

The word lists marked with $_{\textrm{Train}}$ are the ones used to
populate the templates.  The lists marked with $_{\textrm{Test}}$
are the ones used to learn a subspace in the embeddings.  These two
types do not intersect, and when that subscript is omitted, it
implies that it is the union of the two lists -- the full list.
These full lists are used to assess the stability of the associated
subspaces by considering the principal values.

\paragraph{\textsf{Adherents}$_{\textrm{Train}}$}
~\\

\begin{tabular}{p{0.9\linewidth}}
	atheist, 
	baptist, 
	catholic, 
	christian, 
	hindu, 
	methodist, 
	protestant, 
	shia
\end{tabular}

\paragraph{\textsf{Adherents}$_{\textrm{Test}}$}
~\\

\begin{tabular}{p{0.9\linewidth}}
	adventist, 
	anabaptist, 
	anglican, 
	buddhist, 
	confucian, 
	jain, 
	jew, 
	lutheran, 
	mormon, 
	muslim, 
	rastafarian, 
	satanist, 
	scientologist, 
	shinto, 
	sikh, 
	sunni,
	taoist
\end{tabular}

\paragraph{\textsf{Demonyms}$_{\textrm{Train}}$}
~\\

\begin{tabular}{p{0.9\linewidth}}
	american, 
	chinese, 
	egyptian, 
	french, 
	german, 
	korean, 
	pakistani, 
	spanish
\end{tabular}

\paragraph{\textsf{Demonyms}$_{\textrm{Test}}$}
~\\

\begin{tabular}{p{0.9\linewidth}}
	belarusian, 
	brazilian, 
	british, 
	canadian, 
	danish, 
	dutch, 
	emirati, 
	georgian, 
	greek, 
	indian, 
	iranian, 
	iraqi, 
	irish, 
	italian, 
	japanese, 
	libyan, 
	moroccan, 
	nigerian, 
	peruvian, 
	qatari, 
	russian, 
	saudi, 
	scottish, 
	swiss, 
	thai, 
	turkish, 
	ukrainian, 
	uzbekistani, 
	vietnamese, 
	welsh, 
	yemeni, 
	zambian
\end{tabular}

\paragraph{\textsf{Countries}}
~\\

\begin{tabular}{p{0.9\linewidth}}
	america, 
	belarus, 
	brazil, 
	britain, 
	canada, 
	china, 
	denmark, 
	egypt, 
	emirates, 
	france, 
	georgia, 
	germany, 
	greece, 
	india, 
	iran, 
	iraq, 
	ireland, 
	italy, 
	japan, 
	korea, 
	libya, 
	morocco, 
	netherlands, 
	nigeria, 
	pakistan, 
	peru, 
	qatar, 
	russia, 
	scotland, 
	spain, 
	switzerland, 
	thailand, 
	turkey, 
	ukraine, 
	uzbekistan, 
	vietnam, 
	wales, 
	yemen, 
	zambia
\end{tabular}

\paragraph{\textsf{Gendered}$_{\textrm{Test}}$}
~\\

\begin{tabular}{p{0.9\linewidth}}
	man, woman,
	guy, girl, 
	gentleman, lady
\end{tabular}

\paragraph{\textsf{Gendered}}
~\\

\begin{tabular}{p{0.9\linewidth}}
	man, woman, 
	himself, herself, 
	john, mary, 
	father, mother,
	boy, girl,
	son, daughter,
	his, her,
	guy, gal,
	male, female, 
\end{tabular}

\paragraph{\textsf{Polarity}}
~\\

\begin{tabular}{p{0.9\linewidth}}
	awful, 
	dishonest, 
	dumb, 
	evil, 
	great, 
	greedy, 
	hateful, 
	honest, 
	humorless, 
	ignorant, 
	intelligent, 
	intolerant, 
	neat, 
	nice, 
	professional, 
	rude, 
	smart, 
	strong, 
	stupid, 
	terrible, 
	terrible, 
	ugly, 
	unclean, 
	unprofessional, 
	weak, 
	wise
\end{tabular}

\paragraph{\textsf{Verbs}}
~\\

\begin{tabular}{p{0.9\linewidth}}
	ate, 
	befriended, 
	bought, 
	budgeted for, 
	called, 
	can afford, 
	consumed, 
	cooked, 
	crashed, 
	donated, 
	drove, 
	finished, 
	hated, 
	identified, 
	interrupted, 
	liked, 
	loved, 
	met, 
	owns, 
	paid for, 
	prepared, 
	saved, 
	sold, 
	spoke to, 
	swapped, 
	traded, 
	visited
\end{tabular}

\paragraph{\textsf{Objects}}
~\\

\begin{tabular}{p{0.9\linewidth}}
	apple, 
	apron, 
	armchair, 
	auto, 
	bagel, 
	banana, 
	bed, 
	bench, 
	beret, 
	blender, 
	blouse, 
	bookshelf, 
	breakfast, 
	brownie, 
	buffalo, 
	burger, 
	bus, 
	cabinet, 
	cake, 
	calculator, 
	calf, 
	camera, 
	cap, 
	cape, 
	car, 
	cart, 
	cat, 
	chair, 
	chicken, 
	clock, 
	coat, 
	computer, 
	costume, 
	cot, 
	couch, 
	cow, 
	cupboard, 
	dinner, 
	dog, 
	donkey, 
	donut, 
	dress, 
	dresser, 
	duck, 
	goat, 
	headphones, 
	heater, 
	helmet, 
	hen, 
	horse, 
	jacket, 
	jeep, 
	lamb, 
	lamp, 
	lantern, 
	laptop, 
	lunch, 
	mango, 
	meal, 
	muffin, 
	mule, 
	oven, 
	ox, 
	pancake, 
	peach, 
	phone, 
	pig, 
	pizza, 
	potato, 
	printer, 
	pudding, 
	rabbit, 
	radio, 
	recliner, 
	refrigerator, 
	ring, 
	roll, 
	rug, 
	salad, 
	sandwich, 
	shirt, 
	shoe, 
	sofa, 
	soup, 
	stapler, 
	SUV, 
	table, 
	television, 
	toaster, 
	train, 
	tux, 
	TV, 
	van, 
	wagon, 
	watch
\end{tabular}

\paragraph{\textsf{Person Hyponyms}}
~\\

\begin{tabular}{p{0.9\linewidth}}
	acquaintance,
	admirer,
	adolescent,
	adult,
	ancestor,
	clan,
	cohort,
	combatant,
	crew,
	customer,
	employee,
	fellow,
	grown-up,
	in-law,
	neighbor,
	relative,
	resident,
	retiree,
	senior,
	stranger,
	teenager,
	urchin,
	youngster
\end{tabular}

\newpage
\paragraph{\textsf{Occupations}}
~\\

\begin{tabular}{p{0.9\linewidth}}
	accountant, 
	actuary, 
	administrator, 
	advisor, 
	aide, 
	ambassador, 
	architect, 
	artist, 
	astronaut, 
	astronomer, 
	athlete, 
	attendant, 
	attorney, 
	author, 
	babysitter, 
	baker, 
	banker, 
	biologist, 
	broker, 
	builder, 
	butcher, 
	butler, 
	captain, 
	cardiologist, 
	caregiver, 
	carpenter, 
	cashier, 
	caterer, 
	chauffeur, 
	chef, 
	chemist, 
	clerk, 
	coach, 
	contractor, 
	cook, 
	cop, 
	cryptographer, 
	dancer, 
	dentist, 
	detective, 
	dictator, 
	director, 
	doctor, 
	driver, 
	ecologist, 
	economist, 
	editor, 
	educator, 
	electrician, 
	engineer, 
	entrepreneur, 
	executive, 
	farmer, 
	financier, 
	firefighter, 
	gardener, 
	general, 
	geneticist, 
	geologist, 
	golfer, 
	governor, 
	grocer, 
	guard, 
	hairdresser, 
	housekeeper, 
	hunter, 
	inspector, 
	instructor, 
	intern, 
	interpreter, 
	inventor, 
	investigator, 
	janitor, 
	jester, 
	journalist, 
	judge, 
	laborer, 
	landlord, 
	lawyer, 
	lecturer,
	librarian, 
	lifeguard, 
	linguist, 
	lobbyist, 
	magician, 
	manager, 
	manufacturer, 
	marine, 
	marketer, 
	mason, 
	mathematician, 
	mayor, 
	mechanic, 
	messenger, 
	miner, 
	model, 
	musician, 
	novelist, 
	nurse, 
	official, 
	operator, 
	optician, 
	painter, 
	paralegal, 
	pathologist, 
	pediatrician, 
	pharmacist, 
	philosopher, 
	photographer, 
	physician, 
	physicist, 
	pianist, 
	pilot, 
	plumber, 
	poet, 
	politician, 
	postmaster,
	president, 
	principal, 
	producer, 
	professor, 
	programmer, 
	psychiatrist, 
	psychologist, 
	publisher, 
	radiologist, 
	receptionist, 
	reporter, 
	representative, 
	researcher, 
	retailer, 
	sailor, 
	salesperson, 
	scholar, 
	scientist, 
	secretary, 
	senator, 
	sheriff, 
	singer, 
	soldier, 
	spy, 
	statistician, 
	stockbroker, 
	supervisor, 
	surgeon, 
	surveyor, 
	tailor, 
	teacher, 
	technician, 
	trader, 
	translator, 
	tutor, 
	undertaker, 
	valet, 
	veterinarian, 
	violinist, 
	warden, 
	warrior, 
	watchmaker, 
	writer, 
	zookeeper, 
	zoologist
\end{tabular}

\paragraph{\textsf{Rulers}}
~\\

\begin{tabular}{p{0.9\linewidth}}
	administrator,
	admiral,
	aristocrat,
	autocrat,
	bishop,
	boss,
	brass,
	captain,
	chairperson,
	chief,
	chieftain,
	colonel,
	commandant,
	commander,
	commodore,
	consul,
	controller,
	dean,
	despot,
	dictator,
	director,
	don,
	earl,
	elder,
	eminence,
	emir,
	executive,
	general,
	governor,
	imperator,
	judge,
	knight,
	leader,
	manager,
	master,
	mayor,
	monarch,
	noble,
	officer,
	oligarch,
	overlord,
	owner,
	pilot,
	pope,
	premier,
	president,
	priest,
	principal,
	provost,
	regent,
	representative,
	ruler,
	senator,
	shah,
	sheik,
	skipper,
	sovereign,
	sultan,
	superintendent,
	supervisor,
	swami,
	tycoon,
	tyrant,
	vice-president,
	VIP,
	vizier
\end{tabular}

\newpage

\noindent The set of `objects'  used in the template generation includes words from the set objects here, as well as the set of rulers and the set of person hyponyms.

\paragraph{On reproducibility.}
Our code is deterministic and the results in the paper should be reproducible.
We froze all random seeds in code except those deeply buried in learning libraries.
In our preliminary experiments, we found the models are only slightly volatile against this randomness.
With different random runs, the difference in testing accuracies are often in range $(-0.3,0.3)$ on a 100 point scale.
Thus, we believe our result is reproducible offline even though there might be subtle variation.

\end{document}